\documentclass{article}

\pdfoutput=1
\usepackage{microtype}
\usepackage{graphicx}
\usepackage{subfigure}
\usepackage{booktabs} 
\usepackage{natbib}
\usepackage{soul} 

\defcitealias{popperLogicScientificDiscovery2002}{1959/2002}
\defcitealias{tmlrScope}{TMLR, n.d.}
\defcitealias{tmlrHome}{TMLR (n.d.)}
\defcitealias{dmlr}{DMLR, n.d.}
\defcitealias{neurips}{NeurIPS, n.d.}
\defcitealias{icbinbRepo}{ICBINB Initiative, n.d.}
\defcitealias{reproChallenge}{Sinha et al., 2023}

\usepackage[hyphens]{url}
\usepackage{hyperref}
\usepackage{doi} 

\newcommand{\glos}{\ensuremath{^*}}

\usepackage[accepted]{icml2024}

\usepackage{amsmath}
\usepackage{amssymb}
\usepackage{mathtools}
\usepackage{amsthm}
\usepackage{enumitem}

\usepackage[capitalize,noabbrev]{cleveref}

\theoremstyle{plain}

\theoremstyle{definition}

\theoremstyle{remark}

\usepackage[textsize=tiny]{todonotes}

\icmltitlerunning{Rethinking Empirical Research in Machine Learning}

\begin{document}

\twocolumn[
\icmltitle{Position: Why We Must Rethink Empirical Research in Machine Learning
}



\icmlsetsymbol{equal}{*}

\begin{icmlauthorlist}
\icmlauthor{Moritz Herrmann}{ibe,mcml}
\icmlauthor{F. Julian D. Lange}{ibe,mcml}
\icmlauthor{Katharina Eggensperger}{tub}
\icmlauthor{Giuseppe Casalicchio}{stats,mcml}
\icmlauthor{Marcel Wever}{inf,mcml}
\icmlauthor{Matthias Feurer}{stats,mcml}
\icmlauthor{David Rügamer}{stats,mcml}
\icmlauthor{Eyke Hüllermeier}{inf,mcml}
\icmlauthor{Anne-Laure Boulesteix}{ibe,mcml}
\icmlauthor{Bernd Bischl}{stats,mcml}
\end{icmlauthorlist}

\icmlaffiliation{stats}{Department of Statistics, LMU Munich, Munich, 
Germany}
\icmlaffiliation{inf}{Institute of Informatics, LMU Munich, Munich,
Germany}
\icmlaffiliation{ibe}{Institute for Medical Information Processing, Biometry, and Epidemiology, Faculty of Medicine, LMU Munich,
Munich,
Germany}
\icmlaffiliation{tub}{University of Tübingen, Tübingen, Germany}
\icmlaffiliation{mcml}{Munich Center for Machine Learning (MCML), Munich, Germany}

\icmlcorrespondingauthor{Moritz Herrmann}{moritz.herrmann@lmu.de}

\icmlkeywords{Machine Learning, ICML}

\vskip 0.3in
]



\printAffiliationsAndNotice{}  

\begin{abstract}
We warn against a common but incomplete understanding of empirical research in machine learning that leads to non-replicable results, makes findings unreliable, and threatens to undermine progress in the field. To overcome this alarming situation, we call for more awareness of the plurality of ways of gaining knowledge experimentally but also of some epistemic limitations. In particular, we argue most current empirical machine learning research is fashioned as confirmatory research while it should rather be considered exploratory.
\end{abstract}

\section{The Non-Replicable ML Research Enigma}\label{sec:intro}
In his Caltech commencement address \lq\lq Cargo Cult Science''\glos,
\footnote{As our paper contains some jargon, we have included a glossary in the appendix; asterisks $(*)$ in the text denote covered terms.}
Richard Feynman \citeyearpar{feynmanCargoCultScience1974} described how researchers employ practices that conflict with scientific principles to adhere to a certain way of doing things.
\textbf{\textit{This position paper warns against similar tendencies in 
empirical research in machine learning (ML) 
and calls for a mindset change to address methodological and epistemic challenges of experimentation.}}\\[0.75ex]
\textbf{There is ML research that does not replicate.}\hspace{0.25cm}
From an empirical scientific perspective, non-replicable research is a fundamental problem. As Karl Popper \citepalias[p.~66]{popperLogicScientificDiscovery2002} phrased it: \lq\lq non-reproducible single occurrences are of no significance to science."\footnote{Reproducible here does not refer to exact \emph{computational reproducibility}\glos\ but generally to arriving at the same scientific conclusions, termed replicability\glos\ in this paper.}
Consequently, ML research that does not replicate has far-reaching epistemic\glos\ \emph{and} practical consequences. From an epistemological\glos\ point of view, it means that research results are unreliable and, to some extent, it calls into question progress in the field. 
In practice, it may jeopardize applied empirical researchers' confidence in experimental results and discourage them from applying ML methods, even though these novel approaches might be beneficial. 
For example, ML is increasingly being used in the medical domain, and this is often promising in terms of patient benefit. 
However, there are also examples indicating that applied researchers (are starting to) have concerns about ML being used in this high-stakes area. Consider, for example, 
this quite drastic warning by \citet[p.~2]{dhiman2022risk}:
\lq\lq Machine learning is often portrayed as offering many advantages [...].
However, these advantages have not yet materialised into patient benefit [...].
Given the increasing concern about the \emph{methodological quality and risk of bias of prediction model studies} [emphasis added], caution is warranted and the lack of uptake of models in medical practice is not surprising." That is, if the ML community does not improve rigor in empirical methodological research, we think there may be a risk of a backlash against the use of ML in practice. \\
In general, there is a growing body of empirical evidence showing that conclusions drawn from experimental results in ML were overly optimistic at the time of publication and could not be replicated in subsequent studies.
For example, \citet[p.~1]{melisSOTA2018} \lq\lq arrive at the somewhat surprising conclusion that standard LSTM architectures, when properly regularized, outperform more recent models";
\citet[p.~3213]{hendersonDeepReinforcementLearning2018} found for deep reinforcement learning \lq\lq that both intrinsic (e.g. random seeds, environment properties) and extrinsic sources (e.g. hyperparameters, codebases) of non-determinism can contribute to difficulties in reproducing baseline algorithms";
\citet[p.~12]{christodoulou2019systematic} found in a systematic review \lq\lq no performance benefit of machine learning
over logistic regression for clinical prediction models"; \citet[p.~1]{elor2022smote} found in their study on data balancing in classification \lq\lq that balancing does not improve prediction performance for the strong" classifiers; 
see also 
\citet{lucicAreGANsCreated2018}, 
\citet{riquelme2018deep}, 
\citet{raffIndRepro2019},
\citet{herrmannLargeScale2020}, 
\citet{ferrariTroublingRecommender2021}, 
\citet{marieScientificCredibilityMachine2021}, \citet{buchkaOptimisticPerformanceEvaluation2021},
\citet{narang2021transformer},
\citet{goorbergh2022classimbalance}, 
\citet{mateusImageCNN2023}, 
\citet{mcelfresh2023neural}, or the surveys by \citet{liao2021are} and \citet{kapoor2023leakage} for similar findings.
In concrete terms, there is published ML research that is, as Popper would say, \emph{of no significance to science}, but we do not know how much!\\
\textbf{We have been warned; don't we listen?}\hspace{0.25cm}
We are by no means the first to raise these and related issues, and the very fact that we are not the first is a matter of even graver concern. 
We think that empirical research in ML finds itself in a situation where practicing questionable research practices, such as \emph{state-of-the-art-hacking} (SOTA-hacking; \citealp{gencoglu2019harkingdl, hullman2022worst}), has sometimes become more rewarding than following the long line of literature warning against it. \citeauthor{langleyMachineLearningExperimental1988} wrote an editorial \lq\lq Machine Learning as an Experimental Science'' as early as 1988,
and \citeauthor{drummondMachineLearningExperimental2006}
and \citeauthor{handClassifierTechnologyIllusion2006} pointed out problems with experimental method comparison in ML already in 2006. Apart from these specific examples, there is a range of literature over the last decades dealing with similar issues \citep[e.g.,][]{hooker1995testing, mcgeochExpAnalysis2002, johnson2002theoretician, drummondReplicabilityNotReproducibility2009, drummondWarningStatisticalBenchmarking2010, mannarswamyEvolvingAIResearch2018, sculley2018winner's, lipton2018troubling,
bouthillier2019unreproducible, 
liao2021are, 
damour2022underspecification, 
raff2022siren,
lonesHowAvoidMachine2023, trostenQuestionablePracticesMethodological2023}. 
Specifically relevant is the paper by \citet[p.~2]{nakkiran2022incentivizing}, in which they note a \lq\lq perceived lack of legitimacy and real lack of community for good experimental science" (still) exists. 
If we continue not to take these warnings seriously
the amount of non-replicable research will only continue to increase, as the cited \emph{very recent} empirical findings indicate.
\\ 
We do not believe that deliberate actions on the part of individuals have led to this situation but that there is a general unawareness of the fact that, while \lq\lq follow[ing] all the apparent precepts and forms of scientific investigation [in ML]," one can be \lq\lq missing something essential." In particular, this includes that \lq\lq if you’re doing an experiment, you should report everything that you think might make it invalid—not only what you think is right about it: other causes that could possibly explain your results; and things you thought of that you’ve eliminated by some other experiment, and how they worked" \citep[quotes from][p.~11]{feynmanCargoCultScience1974}. 
Misaligned incentives and pressure to publish positive results contribute to this situation \citep[e.g.,][]{
smaldino2016natural}.
\\
\textbf{One of a kind?~At the intersection of formal and empirical sciences.}\hspace{0.25cm}
We believe that one of the main reasons for this is that ML stands, like few other disciplines, at the interface between formal sciences and real-world applications. Because ML has strong foundations in formal sciences such as mathematics, (theoretical) computer science (CS), and mathematical statistics, many ML researchers are accustomed to reasoning mathematically about abstract objects -- ML methods -- using formal proofs. 
On the other hand, ML can also very much be considered a (software) engineering science, to create practical systems that can learn and improve their performance by interacting with their environment.
Lastly, and especially concerning experimentation in ML, 
there exists an applied statistical perspective with a focus on thorough inductive reasoning. With its tradition in data analysis and 
design of experiments, it emphasizes the empirical aspects of ML research.  \\
These different perspectives, with their specific objectives, methodology, and terminology, have their unique virtues, but they also have their blind spots. The formal science perspective aims at providing absolute certainty and deep insights through the definition of abstract concepts and mathematical proofs but is often not well suited to explain complex real-world phenomena, as these concepts and proofs very often have to be based on strongly simplifying assumptions.
The engineering perspective brought us incredible application improvements, 
but at the same time, not all conducted experiments are optimally designed to generalize results beyond the specific application context (which is also often only implicitly defined), as the references provided at the beginning demonstrate.\\
\textbf{A statistical perspective},
which we adopt here, is very sensitive to such empirical issues -- explaining/analyzing real-world phenomena and generalizing beyond a specific context (inductive reasoning) -- and thus particularly suited to explain 1) why ML is faced with non-replicable research, and 2) how a more complete and nuanced understanding of empirical research in ML can help to overcome this situation. With \textbf{empirical ML} we thus mean in a broad sense the systematic investigation of ML algorithms, techniques, and conceptual questions through simulations, experimentation, and observation. It deals with real objects: implementations of algorithms 
-- which are usually more complex
than their theoretical counterparts 
\citep[e.g.,][]{kriegel2017black} --
running on physical computers; data gathered and produced/simulated in the real world; and their interplay. Rather than focusing solely on theoretical analysis and proofs, empirical research emphasizes practical evaluations using real-world and/or synthetic data. Empirical ML, as understood here, requires a mindset very different from engineering and formal sciences and a different approach to methodology to allow for the full incorporation of the uncertainties inherent in dealing with real-world entities in experiments. \\
In our view, the discussed literature, raising similar points, has two main shortcomings:
1) they address only specific aspects of the problem and do not provide a comprehensive picture; 2)
there is a confusion of terminology. For example, \citet{bouthillier2019unreproducible} distinguish between \emph{exploratory} and \emph{empirical} research. 
\citet{nakkiran2022incentivizing} use the term \emph{good experimental research} and contrast it in particular with \emph{improving applications}. 
\citet{sculley2018winner's} talk about \emph{empirical advancements} and  \emph{empirical analysis} that are not complemented by sufficient  \emph{empirical rigor}. And \citet{drummondMachineLearningExperimental2006} discusses ML as an  \emph{experimental science} hardly using the term \emph{empirical} at all.  \\
To overcome these issues, we gather opinions and (empirical) evidence scattered across the literature and different domains and try to develop a \textbf{comprehensive synthesis}. 
For example, similar problems have been discussed in bioinformatics for some time \citep[e.g.,][]{yousefiReportingBiasWhen2010, boulesteixOveroptimismBioinformaticsResearch2010}. We also take into account literature from other, more distant fields facing related issues, such as psychology and medicine. We believe this comprehensive picture will allow for a broader and deeper understanding of the complexity of the situation, which may at first glance appear to be rather easy to solve, e.g., by more (rigorous) statistical machinery or more open research artifacts. It is our conviction that without this deeper understanding, a situation that has been warned about in vain for so long cannot be overcome. 

\section{The Status Quo of Empirical ML}\label{sec:status}
\paragraph{Recent advances.}
It is important to emphasize that there have been encouraging first steps in terms of empirical ML research recently. This includes the newly created publication formats \emph{Transactions on Machine Learning Research}
(TMLR), \emph{Journal of Data-centric Machine Learning Research}
(DMLR), or the NeurIPS \emph{Datasets and Benchmarks Track} launched in 2021.
These venues explicitly include in their scope, e.g., \lq\lq reproducibility studies of previously published results or claims" \citepalias{tmlrScope}, \lq\lq systematic analyses of existing systems on novel datasets or benchmarks that yield important new insight" \citepalias{dmlr}, and \lq\lq systematic analyses of existing systems on novel datasets yielding important new insight" \citepalias{neurips}.
Further examples are 
the \emph{I Can’t Believe It’s Not Better!} (ICBINB) workshop series \citep[e.g.,][]{icbincWS}
and \emph{Repository of Unexpected Negative Results} \citepalias{icbinbRepo}
and efforts towards preregistration
\citep[e.g.,][]{prereg} and reproducibility \citepalias[e.g.,][]{reproChallenge}.
These developments, while very important, are not sufficient in our view to overcome the problems empirical ML faces. For example, while \emph{computational reproducibility}\glos\ may be a necessary condition, it is not a sufficient condition for replicability \citep[e.g.,][]{bouthillier2019unreproducible}.
Furthermore, while the topics in the above formats cover many important aspects of empirical ML, we feel that they do not emphasize enough the importance of true replication of research, which is paramount from an empirical perspective. 
Most importantly, a situation in which a line of research warning us for a long time has been largely neglected will not be overcome by such practical changes alone. 
It also requires a change in awareness -- of the importance of proper empirical ML but maybe even more of its limitations; and that there are different, equally valid types of proper empirical inquiry. We see this lack of awareness evidenced by \citetalias{tmlrHome} itself: \lq\lq TMLR emphasizes technical correctness over subjective significance, to ensure that we facilitate scientific discourse on topics that may \emph{not yet be accepted in mainstream venues} [emphasis added] but may be important in the future." This is expressed in the talk introducing TMLR, too.\footnote{TMLR - A New Open Journal For Machine Learning: \url{https://youtu.be/Uc1r1LfJtds}}
Judging by the example of other empirical sciences, this general lack of awareness of proper empirical ML is certainly the most difficult thing to overcome.
Below we discuss problems we identified as symptoms of this lack.

\textbf{Problem 1: Lack of unbiased experiments and scrutiny.}\hspace{0.25cm}
Most method comparisons are carried out as part of a paper introducing a new method and are usually biased in favor of the new method (see Section~\ref{sec:intro} for examples). \\
\citet[p.~1]{sculley2018winner's} found that \lq\lq [l]ooking over papers from the last year, there seems to be a clear trend of multiple groups finding that prior work in fast moving fields may have missed improvements or key insights due to things as simple as hyperparameter tuning studies$^{[*]}$ or ablation studies."
Moreover, for a neutral method comparison study of survival prediction methods, it has been shown that method rankings can vary considerably depending on design and analysis choices made at the meta-level (e.g., the selected set of datasets, performance metric, aggregation method) and that any method -- even a simple baseline -- can achieve almost any rank 
(\citealp{niessl2022over}; see also \citealp{sonabendCHacking2022}).
We are convinced that it is not far-fetched to conclude that quite often results demonstrating the superiority of a newly proposed method
are obtained by an experimental design favorable to that method. 
\\
As in other disciplines \citep{munafoManifestoReproducibleScience2017}, there are structural issues 
(e.g., publication bias, pressure to publish, lack of replication studies) and questionable practices (e.g., hypothesizing after the results are
known [\citealp[HARKing,][]{kerr1998harking}]
and $p$-hacking [\citealp{simonsohn2014p}])
that contribute to this lack of unbiased experiments and scrutiny. 
At the individual level, in particular, there is a lack of awareness that method comparisons performed as part of a paper introducing a new method are not well suited to draw reliable conclusions about a method beyond the datasets considered, especially if 1) the number of datasets considered is small~\citep{dehghani2021bechmarklottery,koch2021rrr}, 2) there is meta-level overfitting on a single benchmark design~\citep{recht2019imagenet,beyer2020imagenet}, 3) the set of datasets selected for the experiments is biased in favor of the newly proposed method, and 4) the authors are much more familiar with the new method than with its competitors, as is the case frequently \citep[]{johnson2002theoretician, boulesteixPleaNeutralComparison2013, boulesteixEvidencebasedComputationalStatistics2017}. Furthermore, it is very easy to artificially make a method appear superior \citep[e.g.,][]{jelizarowOveroptimismBioinformaticsIllustration2010, norelSelfAssess2011, niessl2022over, ullmann2023over, pawel2024pitfalls, niesslCrossDesign2024}, and publication bias towards positive results is a strong incentive to engage in SOTA-hacking and demonstrate the superiority of a newly proposed method \citep{sculley2018winner's, gencoglu2019harkingdl}. \\
At the system level
there is a publication bias and a lack of replication and neutral method comparison studies \citep[e.g.,][]{boulesteixPleaNeutralComparison2013, boulesteixPubBias2015}. 
\citet[p.~1]{sculley2018winner's} \lq\lq observe that the rate of empirical advancement [larger and more complex experiments] may not have been matched by consistent increase in the level of empirical rigor across the field as a whole."
In unsupervised learning, the problem is more pronounced than in supervised learning because 
\lq\lq there is much less of a benchmarking tradition in the clustering area than in the field of supervised learning" \citep[p.~2; see also \citealp{zimmermannMethodEvaluationParameterization2020}]{mechelen2023whitepaper}. 

\textbf{Problem 2: Lack of legitimacy.} \hspace{0.25cm}
The second problem highlights a specific aspect of the lack of awareness of how different types of empirical research can contribute to ML. The problem was addressed by \citet{nakkiran2022incentivizing} and we completely agree with their description: \\
\hspace{0.25cm}\lq\lq In mainstream ML venues, there is a perceived lack of legitimacy and a real lack of community for good experimental science -- which neither proves a theorem nor improves an application. This effectively suppresses a mode of scientific inquiry which has historically been critical to scientific progress, and which has shown promise in both ML and in CS more generally"
\citep[p.~2]{nakkiran2022incentivizing}. \\
They identify a strong bias of the ML community towards mathematical proofs (formal science perspective) and application improvements (engineering perspective), while \emph{good experimental science} that does not focus on one of the above is not incentivized nor encouraged. \citet{nakkiran2022incentivizing} see this evidenced by the lack of specific subject areas, the exclusion from recent calls for papers, the lack of explicit guidelines for reviewers, and the organization of separate workshops on experimental scientific investigation at major ML conferences. In particular, reviewers \lq\lq often ask for application improvements" and \lq\lq for \lq theoretical justification' for purely experimental papers" \citep[pp.~2--3]{nakkiran2022incentivizing}.
Together these factors point to a structural problem hindering the recognition and promotion of some sorts of experimental research in ML. 
\\
We completely agree with this view but think it may not immediately be clear what distinguishes improving an application from good experimental science at first sight.\footnote{
To avoid misunderstandings: we do consider mathematical proofs and application improvements very valuable research!
} As we understand it, the focus on application improvement means that much empirical/experimental research in ML focuses on developing a new method and demonstrating that it is superior to existing methods by improving some (predictive) performance metric on specific real-world benchmark datasets. 
Good experimental science, on the other hand, is not about improving performance. It is about improving understanding and knowledge of a problem, a (class of) methods, or a phenomenon.
\citet[p.~2]{sculley2018winner's} emphasize that \lq\lq [e]mpirical studies [in ML] have become challenges to be \lq won', rather than a process for developing insight and understanding. Ideally, the benefit of working with real data is to tune and examine the behavior of an algorithm under various sampling distributions, to learn about the strengths and weaknesses of the algorithms, as one would do in controlled studies." And \citet[p.~9]{rendsburgNetGANGANRandom2020} argue,
\lq\lq it is particularly important that our community actively attempts to understand the inherent inductive biases, strengths, and also the weaknesses of algorithms. Finding examples where an algorithm works is important -- but maybe even more important is to understand under which circumstances the algorithm produces misleading results." 

\textbf{Problem 3: Lack of conceptual clarity and operationalization.}\hspace{0.25cm} There is a perceived lack of clarity about some important abstract concepts that are the objects of ML research on the one side and a lack of clear operationalization\glos~in empirical investigations on the other side. Both aspects affect the validity of experiments in empirical ML.\\
This problem is the most complex one and probably for that reason the most difficult to describe in precise terms \citep[cf.][]{saitta1998learning}. However, since we think that this problem affects the validity of empirical research in ML in a fundamental way, an account of empirical ML that does not attempt to make it tangible would be incomplete. We aim to narrow down the problem by explicating examples for supervised learning and unsupervised learning. \\
In other sciences such as psychology and physics, \emph{validity}\glos, the fact that the experimental measurement process actually measures what it is intended to measure, is fundamental. It inevitably depends on a strict and thorough operationalization in what way abstract concepts that are to be measured relate to measurable entities in the real world. Note that \lq\lq[o]perational analysis is an excellent diagnostic tool for revealing where our knowledge is weak, in order to guide our efforts to strengthening it. The Bridgmanian ideal$^{[*]}$ is always to back up concepts with operational definitions, that is, to ensure that every concept is independently measurable in every circumstance under which it is used" \citep[p.~147]{chang2004inventing}. It is puzzling that validity and other quality criteria of empirical research have gained little attention in ML so far \citep[e.g.,][]{myrtveitReliabilityValidityComparative2005, segebarthObjectivityReliabilityValidity2020, rajiAIEverything2021}.\\
\textbf{Experimental validity in supervised learning.}\hspace{0.25cm}
For supervised learning, the problem can be exemplified by the question of inference from experimental results on real data in method comparison and evaluation studies.\footnote{Another example independently affecting validity is underspecification, which \lq\lq is common in modern ML pipelines, such as those based on deep learning" 
\citep[p.~2]{damour2022underspecification}.} 
Typically, the goal is to generalize the observed performance difference between methods to datasets that were not included in a study, 
which would require specifying when datasets are from the same/different domain. The problem is that it is not at all clear in what sense results obtained from one set of real datasets can be generalized to any other set of datasets, as this would require a clear understanding of the distribution of the data-generating processes by which each dataset is generated \citep[e.g.,][]{aha1992generalizing, salzberg1997comparing, boulesteixStatisticalFrameworkHypothesis2015, herrmann2022towards, strobl2024against}. Without a 
definition of the population of data-generating processes, i.e., (some) clarity about an abstract concept, it can be argued that it is not clear what a real data comparison study actually measures. In other words, the collection of datasets considered \lq\lq will not be representative of real data sets in any formal sense" \citep[p.~12]{handClassifierTechnologyIllusion2006}. \citet[p.~4]{dietterichApproximateStatisticalTests1998} even went so far as claiming that how to perform benchmark experiments on real datasets properly is
\lq\lq perhaps the most fundamental and difficult question in machine learning."
\\
\textbf{Experimental validity in unsupervised learning.}\hspace{0.25cm}
Arguably, the situation is even more involved in unsupervised learning \citep[e.g.,][]{kleinbergImpossibilityTheoremClustering2002, luxburgClustering2012, zimekThereBackAgain2018, herrmann2022towards}. 
First of all, “there is no [...] direct measure of success. It is difficult to ascertain the validity of inferences drawn from the output of most unsupervised learning algorithms" \citep[p.~487]{hastieElementsStatisticalLearning2009}.
This is aggravated by an ambiguity about the abstract concepts of interest. Consider, for example, cluster analysis.\footnote{
See also \citet{herrmannOutlierFramwork2023} for outlier detection.} 
Usually, clusters are conceptualized as the modes of a mixture of (normal) distributions. However, there is a different perspective that considers cluster analysis from a topological perspective and conceptualizes clusters as the connected components of a dataset \cite{niyogiTopologicalViewUnsupervised2011}. It is not clear if these different notions of clusters 1) conceptualize clusters equally well, 2) can be related to the same real-world entities, and 3) whether clustering methods developed based on these different notions are equally suitable for all clustering problems. There is some evidence that suggests this is not the case \citep{herrmannTopoClust2023}. 

\textbf{Problem summary.}\hspace{0.25cm} We argue that much empirical ML research is prone to overly optimistic, unreliable, and difficult-to-refute judgments and conclusions. Many experiments in empirical ML research are based on insufficiently operationalized experimental setups, partially due to ambiguous and inconclusive conceptualizations underlying the experiments. 
To draw more reliable conclusions, we need more explicit, context-specific operationalizations and clearer delineations of the abstract concepts that are to be investigated. 
Recall that \lq\lq [o]perational analysis is an excellent diagnostic tool for revealing where our knowledge is weak, in order to guide our efforts in strengthening it" \citep[p.~147]{chang2004inventing}. 
That sometimes good experimental research is not encouraged enough in ML (see Problem 2) and biased experiments still occur more often than desirable (see Problem 1), exacerbates the situation considerably. 
The former is an excellent approach for improving insight and understanding in the sense outlined above, biased experiments tend to make this more difficult. 
These aspects are becoming specifically important in deep learning where the sheer complexity of today's models, especially of foundation models, makes mathematical analysis extremely difficult. Instead, the analysis often needs to be largely experimental and thus requires thorough experimentation at the highest possible level.

\section{Improving the Status Quo: More Richness in Empirical Methodological Research}\label{sec:improve}

\paragraph{A unifying view: We need exploratory and confirmatory.}
Confirmatory research\glos, also known as hypothesis-testing research, aims to test preexisting hypotheses to confirm or refute existing theories. Researchers design specific studies to evaluate hypotheses derived from existing knowledge experimentally. Typically, this involves a structured and predefined research design, a priori hypotheses, and often statistical analyses to draw conclusive inferences. 
In contrast, exploratory research is an open-ended approach that aims to gain insight and understanding in a new or unexplored area. It is often conducted when little is known about the phenomenon under study. 
It involves gathering information, identifying patterns, and formulating specific hypotheses for further investigation. One of our main points is that to improve empirical ML towards more thorough, reliable, and insightful methodological research both exploratory and confirmatory research are needed in ML \citep[cf.][]{tukey1980we}. \\
In general, the problems described can be placed in this broader epistemic context.
We argue that most empirical research in ML is perceived as confirmatory research, when it should rather be considered to be exploratory from an epistemic perspective \citep[see also][]{bouthillier2019unreproducible}. 
At the same time, purely exploratory methodological research focusing on improving insight and understanding experimentally \citep[cf.][]{dietterich1990exploratory} and research like neutral method comparison and replication studies, which can be considered more rigorous in the confirmatory sense, are not seen as an equally important contribution to the field. \\
For the time being, it is worth making this distinction, 
yet, we discuss why it is an oversimplification from an epistemic perspective in Section~\ref{sec:beyond} -- even more so, because we distinguish two types of exploratory empirical methodological research in the following:\footnote{It is not our intention to establish a precise terminology, but we think this structure will be of assistance to the reader.} \emph{insight-oriented exploratory research}\glos\ in contrast to \emph{method-developing exploratory research}\glos. 
We think insight-oriented exploratory research is what \citet{nakkiran2022incentivizing} mean by good experimental research, and what they mean by application improvements is a conflation of both method-developing exploratory and (supposedly) confirmatory research. 
\newpage

\textbf{More insight-oriented exploratory research.}\hspace{0.25cm}
In principle, the good thing about moving towards more insight-oriented exploratory methodological research in ML is that there are no epistemological obstacles to overcome. The neighboring field \textit{data mining and knowledge discovery} clearly has an exploratory nature and is very much in the spirit of Tukey's exploratory data analysis. There are also already some examples of influential ML research that can be considered insight-oriented and exploratory, e.g., \citet{frankle2019lottery}, \citet{belkin2019reconciling}, \citet{recht2019imagenet}, \citet{rendsburgNetGANGANRandom2020}, \citet{zhang2021rethinking}, or \citet{power2021grokking}. 
So, rather than epistemic aspects, it is the incentives and attitudes in scientific practice towards this type of research that are an obstacle to its successful dissemination. 
In particular, an alleged lack of novelty and originality is often invoked, which leads to rejections. Yet, without the esteem expressed by acceptance for publication, in particular in major ML venues, there is simply little incentive to engage in exploratory ML research. 
More importantly, it reinforces the impression among students and young scientists that exploratory research is not an integral part of science. 
It is therefore necessary to stimulate, encourage, and provide opportunities to make such research visible. \citet[pp.~4--5]{nakkiran2022incentivizing} propose to establish a special subject area within ML conferences for \lq\lq Experimental Science of Machine Learning,'' focusing on \lq\lq experimental investigation into the nature of learning and learning systems." 
The types of papers outlined include those with \lq\lq surprising experiments," \lq\lq empirical conjectures," \lq\lq refining existing phenomena," \lq\lq formalizing intuition," and presentation of \lq\lq new measurement tool[s]," all aiming to improve the understanding of ML empirically. 
They also provide guidelines specifically tailored to the review of this type of research.

\textbf{More (actual) confirmatory research.}\hspace{0.25cm}
As outlined, we believe most current empirical ML research (i.e., application improvements) is a mixture of method-developing exploratory research and (supposedly) confirmatory research.\footnote{In a sense, this limits the potential of the former and renders the latter largely useless, with biased experiments as a result.}
For this reason, 
we add a focus on well-designed, neutral method comparison and replication studies. 
The scrutiny and rigor these examples of (actual) confirmatory empirical research provide are sorely needed if we are to work toward more reliable and replicable research. \\
\textbf{Neutral method comparison studies} include experiments that are less biased in favor of newly proposed methods \citep{boulesteixPleaNeutralComparison2013, lim2000comparison, aliAloSelect2006, fernandez2014we}. 
First, this includes prespecified, strictly adhered-to designs of the experimental setup, including in particular a clearly specified set of datasets and tasks. Ideally, neutral comparison studies focus on the comparison of \emph{already existing} methods and are carried out by a group of authors approximately equally familiar with all the methods under consideration \cite{boulesteixPleaNeutralComparison2013}. 
Such studies ensure more neutrality
and are less prone to overly optimistic conclusions than studies proposing a method, since there is much less of an incentive to promote a particular method.
Second, proper \textbf{uncertainty quantification} is required when analyzing empirical results in ML, especially w.r.t. the different stages of inference (e.g., model fitting, model selection, pipeline construction, and performance estimation)
\citep[see][]{nadeauInferenceGeneralizationError2003, bengioNoUnbiasedEstimator2004, hothornDesignAnalysisBenchmark2005, batesCrossvalidationWhatDoes2023}.
Moreover, if \textbf{statistical significance testing}
is to be conducted to test for \emph{statistically} significant performance differences across different real-world datasets, as described, e.g., by \citet{demsarStatisticalComparisonsClassifiers2006}, \citet{eugsterDomainBasedBenchmarkExperiments2012}, \citet{boulesteixStatisticalFrameworkHypothesis2015}, or \citet{eisingaExactPvaluesPairwise2017}, the methodological rigor established in other empirical domains should be applied \cite{munafoManifestoReproducibleScience2017}, in particular, efforts towards prior sample size calculations are important \cite{boulesteixEvidencebasedComputationalStatistics2017}. \\%
Moreover, we need more \textbf{replication studies} and \textbf{meta-studies}. These types of research face similar reservations as insight-oriented exploratory experimental research. However, replication studies are indispensable to assess the amount of non-replicable research and to prevent it from being increased. 
Such studies attempt to reach the same scientific conclusions as previous studies, to provide additional empirical evidence for observed phenomena. Meta-studies analyze and summarize the so accumulated evidence on a specific phenomenon. This process is the default to reach conclusions in other sciences and is important as single studies can be false and/or contradict each other.
In ML, this can range from studies that attempt to replicate an experiment exactly \citep[e.g.,][]{lohmannRepSimStud2022} or slightly modify an experiment's design (e.g., by using a different set of data in the replication of a neutral comparison study) to more comprehensive tuning and ablation studies of experiments conducted in method-developing research \citep[e.g.,][]{rendsburgNetGANGANRandom2020, kobak2021initialization}. The latter certainly overlaps with insight-oriented exploratory research. 
It is important to emphasize that it is in the nature of things that a replication is not an original or novel scientific contribution in the conventional sense, and not necessarily can important new insights be gained beyond the replication of previously observed results. Rather, it is an explicit attempt to arrive at the same results and conclusions as a previous study. The scientific relevance, which is well acknowledged in other empirical sciences such as physics or medicine, lies in gathering additional empirical evidence for a hypothesis through a successful replication. 
Moreover, a replication study may, but does not necessarily, also raise epistemic questions, point to experimental improvements, or provide refined concepts, especially in failed replications. 
\newpage
\textbf{More infrastructure.}\hspace{0.25cm}
To achieve this, practical limitations also need to be overcome. We require more dedicated infrastructure to make the proposed forms of research more (easily) realizable. In particular, there is a need for more and better open databases of well-curated and well-understood datasets such as OpenML~\citep{vanschorenOpenML2013} or OpenML Benchmarking Suites~\citep{bischl2021benchmarking}. Moreover, well-maintained open-source software 
for systematic benchmark experiments, such as the AutoML Benchmark~\citep{gijsbers2024amlb}, HPOBench~\citep{eggensperger2021hpobench}, NAS-Bench-Suite~\citep{mehta2022nasbenchsuite}, or AlgoPerf~\citep{dahl2023benchmarking}, are needed. Platforms for public leaderboards and model sharing (e.g., Hugging Face) are another important aspect, although some of these platforms are geared towards horse racing based on predictive performance and therefore do not necessarily also provide scientific insights or interpretability. Yet, the standards and automatic nature of such platforms have the advantage that they offer concrete reference points for criticism and debate. 
Finally, reviewer guidelines implementing our suggestions and dedicated venues for currently hard-to-publish empirical work will allow the full potential of empirical ML to be realized \citep{sculley2018winner's, nakkiran2022incentivizing}. \\
Moreover, without \textbf{more education}, none of this will be possible.
Given the different perspectives -- formal science, engineering, statistical -- from which 
ML can be viewed, it is very difficult to include each in the appropriate depth in a single study program. 
While a recent survey of 101 undergraduate data science programs in the U.S. showed that all included an introductory course in statistics
\citep{bilehassanDataScienceCurriculum2021},
statistics has only recently (2023) been included as a 
core topic in the curriculum recommendations for CS\glos\ \cite{acmCSCurricula2023}. 
It is also questionable if introductory courses are sufficient to avoid crucial gaps that can lead to the adoption of questionable research practices \citep[cf.][]{gigerenzer2018rituals}.
Furthermore, nearly no study program contains a dedicated course on \emph{design and analysis of (computer) experiments} \citep{santner2003design, box2005design, dean2017design}, which we deem especially relevant for our context here.
In general, we agree with \citet[pp.~16--17]{deveauxCurriculumGuidelinesUndergraduate2017} that many \lq\lq courses traditionally found in computer science, statistics, and mathematics offerings should be redesigned for the data science [or ML] major in the interest of efficiency and the potential synergy that integrated courses would offer." 

Finally, we would like to offer \textbf{concrete and practicable advice} to specific target groups, in addition to the general recommendations above.\\
\textbf{Advice for junior researchers.} (1) Read the positive examples of insight-oriented exploratory research in ML (listed above), about the design of experiments, the critical discussion on statistical testing, and the basics of philosophy of science. (2) Educate yourself in Open Science practices \citep[e.g., see][]{turingway2023}. (3) Engage with researchers from other disciplines as data (the one on which ML models are trained) can only really be understood if one understands how it was generated. (4) Consider making empirical research in ML a (partial) research focus. \\
\textbf{Advice for senior researchers.} (1) Allow your junior researchers to write (great) papers on empirical aspects of ML, even if those may be relatively difficult to publish in major venues for now. Our personal experience is that these papers can still be highly cited and become very influential. (2) Learn from other fields; what we are experiencing in terms of non-replicable research is not a new phenomenon. (3) Please do not perceive this paper as an attack on ML but rather as an honest attempt to improve it and, more importantly, to improve its impact. \\
\textbf{Advice for venue organizers and editors} \citep[see also][]{nakkiran2022incentivizing}. (1) Encourage all forms of proper empirical ML to be submitted (in particular, this includes insight-oriented exploratory research), e.g., by creating special tracks or adding keywords but also by allowing such work on main tracks. The idea is to create special measures for the topic to increase awareness but not to isolate or ban all such papers to special (workshop) tracks with (potentially) lower perceived value. (2) Consider giving out awards for positive examples of these types of research. (3) Consider establishing positions like \emph{reproducibility} and \emph{replicability} editors for venues and journals. (4) Give concrete advice for best practices, so authors and reviewers have clear guidelines to follow. Note that this should not be confined to asking \lq\lq Were the empirical results subjected to statistical tests?'' (without further information); this is close to the opposite of what we think is needed.\footnote{One anonymous reviewer also suggested that venues start collecting metadata on reasons for rejection. Such data could serve as a basis to evaluate if certain types of ML research face a systematic bias.}

\section{Beyond the Status Quo: Rethinking Empirical ML as a Maturing Science}\label{sec:beyond}
\paragraph{The exploratory-confirmatory research continuum.}
With ML's strong foundation in formal sciences, where absolute certainty can be achieved by formal proofs, the clear distinction between exploratory and confirmatory research that has been invoked so far may seem natural. Yet, from an empirical perspective, i.e., whenever one deals with entities in the real world, it is itself an oversimplifying dichotomy, and empirical research is better thought of as a continuum from exploratory to confirmatory, with an ideal of purely exploratory research at one end and of strictly confirmatory research at the other \citep[e.g.,][]{Wagenmakers2012agenda, oberauer2019addressing, szollosi2021arrested, Scheel2021why, devezer2021case, rubin2022exploratory, hofler2022means, fife2022understanding}.\\
Based on that notion, \citet{fife2022understanding} argue that \lq\lq psychology may not be mature enough to justify confirmatory research" (p.~453)
and that \lq\lq[t]he maturity of any science puts a cap on the exploratory/confirmatory continuum" (p.~462).
Given the similarities between research in psychology and ML as described by \citet{hullman2022worst}, we think similar holds for ML and we suggest ML should be considered as a maturing (empirical) science as well.\footnote{There are also differences between ML and psychology considerably simplifying our lives: we usually do not experiment on humans but algorithms on computers and have more control over experiments, larger sample sizes, and lower experimental costs.}
\citet[p.~355]{hullman2022worst} \lq\lq identify common themes in reform discussions, like overreliance on asymptotic theory and non-credible beliefs about real-world data-generating processes."
That said, confirmatory research in ML as advocated in Section~\ref{sec:improve} is still very different from \emph{strict} confirmatory research in other disciplines. Rather it can be seen as \emph{rough} confirmatory research \citep{fife2022understanding, tukey1973exploratory} that follows the same principles, but -- as outlined -- it is unclear 
how results can be generalized  
(e.g., using statistical tests) -- a cornerstone of strict confirmatory research. But this should not be taken as a caveat. Rough confirmatory research allows for flexibility that strict confirmatory research does not \citep{fife2022understanding}. \\
The framework 
proposed by \citet[p.~1]{heinze2024phases} can be seen as a way of mapping this rather abstract idea into more concrete guidelines for scientific practice.
In the context of biostatistics, they propose to consider four phases of methodological research, analogous to clinical research in drug development: \lq\lq (I) proposing a new methodological idea while providing, for example, logical reasoning or proofs, (II) providing empirical evidence, first in a narrow target setting, then (III) in an extended range of settings and for various outcomes, accompanied by appropriate application examples, and (IV) investigations that establish a method as sufficiently well-understood to know when it is preferred over others and when it is not; that is, its pitfalls." 

\textbf{Statistical significance tests: Words of caution, revisited!}
The problem of empirical research as a continuum is more involved epistemologically and cannot be discussed in full detail here. An important aspect that needs to be discussed is its relation to the misguided use of statistical testing. 
This point has been made by \citet{drummondMachineLearningExperimental2006} before and in more detail. We revisit it here, enriching it with more recent literature on the issue. In particular, \emph{routinely} adding statistical machinery to an (already underspecified and/or biased) experimental design to test for \emph{statistically significant} differences in performance -- as is frequently done and/or explicitly asked for \citep[e.g.,][]{hendersonDeepReinforcementLearning2018, marieScientificCredibilityMachine2021} -- 
does not improve the epistemic relevance of the results by much nor does it add much additional insight over other data aggregations. In fact, \lq\lq [s]tatistical significance was never meant to imply scientific importance," and you should not \lq\lq conclude anything about scientific or practical importance based on statistical significance (or lack thereof)" \citep[pp.~2, 1]{wasserstein2019moving}. On the contrary, the misguided beliefs in and use of \emph{statistical rituals} \citep{gigerenzer2018rituals} is largely responsible for the replication crisis in other empirical disciplines.\\
The reasons are complex. First of all, the modern theory of statistical hypothesis testing (SHT) is a conflation of \emph{two historically distinct types of testing theory}\glos.
Important epistemological questions about when statistical tests are appropriate are obscured by this mixed theory \citep[e.g.,][]{schneider2015null, gigerenzer2015surrogate, rubin2020repeated}. 
More importantly, specifically for experiments in ML, both theories are developed for experimental designs based on samples randomly drawn from a population of interest \citep{schneider2015null}. 
In general, the assumptions underlying the theory of statistical testing as an inferential tool 
are usually not met in many applications \citep{greenland2023pvals}. 
In fact, the editors of the \emph{The American Statistician} special issue \lq\lq Statistical Inference in the 21st Century: A World Beyond $p < 0.05$''
went so far as to conclude, \lq\lq based on [their] review of the articles in this special issue and the broader literature, that it is time to stop using the term \lq statistically significant' entirely" \citep[p.~2]{wasserstein2019moving}.
\\ 
Note that we want to warn against an \emph{overemphasis} on as well as an \emph{uncritical} use of statistical tests; we do not argue against statistical testing in general.
Quite the contrary, we argue for a more diverse set of analysis tools (applied with care and critical reflection), including but not limited to statistical testing. We also want to stress that statistical testing cannot remedy more fundamental problems such as poor experimental design. To summarize the main points, we emphasize:
\begin{itemize}[itemsep=1pt,topsep=0pt] 
    \item Valid statistical testing inevitably depends on a thorough and well-designed experimental setup.
    \item Statistical testing should not be applied routinely and requires thought and careful preparation to be valid and insightful.
    \item Improper statistical testing and/or its uneducated interpretation are – widely acknowledged – a main driver for non-replicable results in other empirical sciences.
    \item The discussion about these issues has been going on for decades and has resulted in a large body of literature, some of which is condensed in the mentioned special issue of \emph{The American Statistician}.
\end{itemize}
 
So, while we argue for more experiments in 
a confirmatory spirit to improve the status quo of empirical ML (see Section \ref{sec:improve}), especially using neutral method comparison and replication studies, we also emphasize that it is important to keep in mind their current epistemic limitations. In particular, we warn against common misconceptions about and inappropriate use of SHT.
The problem is that the underlying \lq\lq misunderstandings stem from a set of interrelated cognitive biases that reflect innate human compulsions which even the most advanced mathematical training seems to do nothing to staunch, and may even aggravate: Dichotomania, the tendency to reduce quantitative scales to dichotomies;
nullism, the tendency to believe or at least act as if an unrefuted null hypothesis is true; 
and statistical reification, the tendency to forget that mathematical arguments say nothing about reality except to the extent the assumptions they make (which are often implicit) can be mapped into reality in a way that makes them all correct simultaneously" \citep[p.~911]{greenland2023pvals}.

\textbf{Most current empirical ML research should rather be viewed as exploratory.}\hspace{0.25cm}
As outlined, confirmatory research aims to test preexisting hypotheses, while exploratory research  
involves gathering information, identifying patterns, and formulating specific hypotheses for further investigation.
We think, currently, most of the empirical research in ML is conducted as part of a paper introducing a new method and is \emph{fashioned} as confirmatory research even though it is exploratory in nature. In our view, this is reflected especially in the routine use of statistical tests to aggregate benchmark results: the exploratory phase of method development (e.g., trying out different method variants) largely invalidates post hoc statistical tests. As \citet[p.~2]{strobl2024against} put it: \lq\lq In methodological research, comparison studies are often published either with the explicit or implicit aim to promote a new method by means of showing that it outperforms existing methods." In other words, the conducted experiments are set up to \emph{confirm} the (implicit) hypothesis that the proposed method
constitutes an improvement.
Systemic pressures and 
conventions, as well as ML's strong roots in formal sciences and focus on improving applications, encourage this mindset and the practice of invoking confirmatory arguments. This is
expressed in statements such as that \lq\lq [i]t is well-known that reviewers ask for application improvements" and \lq\lq for 
\lq theoretical justification' for purely experimental papers, even when the experiments alone constitute a valid scientific contribution"
\citep[pp.~2--3]{nakkiran2022incentivizing}.
The problem with not emphasizing the exploratory nature is that \lq\lq exploratory findings have a slippery way of \lq transforming’ into planned findings as the research process progresses” \citep[p.~275]{calinjageman2019newstatistics} and \lq\lq[a]t the bottom of that slippery slope one often finds results that don’t reproduce" \citep[p.~3]{wasserstein2019moving}. 
Shifting the focus to an exploratory notion of method development is an opportunity to fully allow \lq\lq to understand under which circumstances the algorithm produces misleading results" \citep[p.~9]{rendsburgNetGANGANRandom2020} and to \lq\lq learn about [its] strengths and weaknesses" \citep[p.~2]{sculley2018winner's} and clearly report them.

\section{Conclusion}\label{sec:conclusion}
This work offers perspectives on ML that outline how it should move from a field being largely driven by mathematical proofs and application improvements to \emph{also} becoming a full-fledged empirical field 
driven by multiple types of 
experimental research. By providing concrete practical guidance but at the same time moderating expectations of what empirical research can achieve, we wish to contribute to greater overall reliability and 
trustworthiness. \\
\textbf{For every don't, there is a do.}\hspace{0.25cm}
However, we are aware that our explanations may initially leave the reader unsatisfied when it comes to translating the conclusions into scientific practice. For example, those who were hoping for guidelines on the correct use of statistical tests may well be at a complete loss. However, we do not believe that this is actually the case. If you are inclined to perform statistical tests as described
by \citet{demsarStatisticalComparisonsClassifiers2006}, do so, but also be aware of the Do-lists described by \citet[Ch. 3, 7]{wasserstein2019moving}. 
In this regard, we consider the following comment by \citeauthor{wasserstein2019moving} (ib., p.~6) very noteworthy: \lq\lq Researchers of any ilk may rarely advertise their personal modesty. Yet, the most successful ones cultivate a practice of being modest throughout their research, by understanding and clearly expressing the limitations of their work." Furthermore, do not only rely on real data, use simulated data as well. Simulations are an excellent tool for operationalization, i.e., mapping abstract concepts to measurable entities. 
Yet, the most important point
is that we should be open to different ways of doing experimental research and should not penalize research
just because it does not follow certain
established conventions. 
As \citet[p.~6]{nakkiran2022incentivizing} put it:
\lq\lq Each paper must be evaluated on an individual basis"; 
this is challenging, but they suggest guidelines. The community should take them up to address this issue.\\ 
\textbf{Embracing inconclusiveness.}\hspace{0.25cm}
Summarizing the 
perspectives on empirical 
ML covered here, and returning to the idea of mature sciences, we believe that for ML to mature as an (empirical) science, a greater awareness of some epistemic limitations,
but also of the plurality of ways to gain insights,
might be all it needs.
We believe that if empirical research is one thing, it is not conclusive and no single empirical study can prove anything with absolute certainty. It must be scrutinized, repeated, and reassessed in a sense of \emph{epistemic iteration}$^*$ \citep{chang2004inventing}.
That said, we conclude by quoting \citeauthor{chang2004inventing}’s thoughts (ib., p.~243) on science in general:
\lq\lq If something is actually uncertain, our knowledge is superior if it is accompanied by an appropriate degree of doubt rather than blind faith. If the reasons we have for a certain belief are inconclusive, being aware of the inconclusiveness prepares us better for the possibility that other reasons may emerge to overturn our belief. With a critical awareness of uncertainty and inconclusiveness, our knowledge reaches a higher level of flexibility and sophistication."

\section*{Impact Statement}
This position paper aims to advance machine learning by addressing practical challenges and epistemic constraints of empirical research that are often overlooked. We believe this has implications for machine learning research in general, as it can help to improve the reliability and credibility of research results. We also believe that our contribution can have a broader positive social and ethical impact by preventing misdirected efforts and resources.

\section*{Acknowledgments}
We thank the four anonymous reviewers for their valuable comments and suggestions.
Katharina Eggensperger is a member of the Machine Learning Cluster of Excellence, EXC number 2064/1 – Project number 390727645. Anne-Laure Boulesteix was partly funded by DFG grant BO3139/9-1.





\bibliographystyle{icml2024}
\bibliography{aa_references}  






\newpage
\appendix
\onecolumn
\section*{Glossary}

\emph{Bridgmanian ideal.} Used by \citet{chang2004inventing} to describe a specific notion of \emph{operationalization}. Refers to Percy Williams Bridgman (1882--1961), Nobel laureate in physics for his work on high-pressure physics, who also made contributions to the philosophy of science. Operational analysis is the topic of his book \emph{The Logic of Modern Physics}, in which he argues in particular that \lq\lq[i]n general, we mean by any concept nothing more than a set of operations; \emph{the concept is synonymous with the corresponding set of operations}" \citep[p.~5
]{bridgman1927logic}.
This strict perspective on operationalization (also referred to as operationalism) has attracted a lot of criticism, see \lq\lq Operationalism'' in \emph{The Stanford Encyclopedia of Philosophy} \citep{changOper2021}. In particular, \citet[p.~148]{chang2004inventing} points out that it builds on \lq\lq an overly restrictive notion of meaning, which comes down to reduction of meaning to measurement, which [Chang] refer[s] to as \emph{Bridgman’s reductive doctrine of meaning}
."

\emph{Cargo Cult Science.} The term cargo cult refers to social movements that originated in Melanesia: \lq\lq The modal cargo cult was an agitation or organised social movement of Melanesian villagers in pursuit of \lq cargo' by means of renewed or invented ritual action that they hoped would induce ancestral spirits or other powerful beings to provide" \citep[p.~1]{lindstromCargoCult2018}. Richard Phillips Feynman (1918--1988), theoretical physicist and Nobel laureate, adapted the term to describe ritualized scientific practices which \lq\lq follow all the apparent precepts and forms of scientific investigation, but [which are] missing something essential" \citep[p.~11]{feynmanCargoCultScience1974}.

\emph{Confirmatory research.} 
Also known as hypothesis-testing research, aims to test preexisting hypotheses to confirm or refute existing theories. Researchers design specific studies to evaluate hypotheses derived from existing knowledge experimentally. Typically, this involves a structured and predefined research design, a priori hypotheses, and often statistical analyses to draw conclusive inferences. It is a well-established term in many fields other than ML. For example, general references are \citet{schwab2020confirm}, \citet{nosek2018preregistration}, and \citet{munafoManifestoReproducibleScience2017}. Field-specific references include \citet{jaeger1998confirm} or \citet{nilsen2020confirm} for biology, \citet{Wagenmakers2012agenda} for psychology, \citet{kimmelman2014distinguishing} for preclinical research, \citet{roettger2021confirm} for linguistics, or \citet{foster2024confirm} for educational research. The term \emph{confirmatory} might appear to be in conflict with the principle of falsification established by Popper \citepalias{popperLogicScientificDiscovery2002}. According to Popper, scientific theories cannot be conclusively confirmed, only falsified. It is important to emphasize that \emph{confirmatory research} has a narrower scope rooted in Neyman-Pearson statistical testing theory (see the glossary entry on \emph{Two historically distinct types of testing theory}). This theory provides a framework for a statistically justified decision between a null hypothesis and an alternative hypothesis based on the available data. The hypothesis to be established (e.g., there is an effect) is usually stated as the alternative hypothesis and confirmation means rejecting the null hypothesis (e.g., there is no effect) for the alternative.

\emph{Curricula recommendations for CS.} The report \emph{Computer Science Curricula 2013}
lists \lq\lq Intelligent Systems'' (including basics in ML) as a Core (Tier2) topic but \lq\lq still believe[s] it is not necessary for all CS programs to require a full course in probability theory [or statistics]" \citep[p.~50]{acmCSCurricula2013}. This has changed with the latest (2023) version insofar as statistics is now considered a CS Core topic in \lq\lq Mathematical and Statistical Foundations'', which is one of several knowledge areas
\cite{acmCSCurricula2023}.

\emph{Epistemic, epistemological.} Both coming from the Greek word for knowledge or understanding, the terms are sometimes used synonymously and sometimes with distinct, more precise meanings. If the distinction is made, \emph{epistemic} relates to knowledge itself, while \emph{epistemological} relates to \lq \lq the study of the nature and grounds of knowledge'' \citep{MerriamWebsterEpistemic}, i.e., epistemology. 
For epistemology, an early edition of \emph{The Stanford Encyclopedia of Philosophy} gives the following definition: \lq\lq Defined narrowly, epistemology is the study of knowledge and justified belief. 
[...] Understood more broadly, epistemology is about issues having to do with the creation and dissemination of knowledge in particular areas of inquiry" \citep{steupEpist2006}. 
The most recent edition states
in more abstract terms that \lq\lq [m]uch recent work in formal epistemology is an attempt to understand how our degrees of confidence are rationally constrained by our evidence [...]" and that \lq\lq epistemology seeks to understand one or another kind of \emph{cognitive success} [...]"
\citep{steupEpist2020}.

\emph{Epistemic iteration.} \citet{chang2004inventing} introduced the concept and defined it in his glossary as a \lq\lq process in which successive stages of knowledge, each building on the preceding one, are created in order to enhance the achievement of certain epistemic goals. It differs crucially from mathematical iteration in that the latter is used to approach a correct answer that is known, or at least in principle knowable, by other means" (p.~253).  For thorough discussions, see Chapters 1 (pp.~46--48) and 5.

\emph{Exploratory research}. As also specified in the main body of the paper, refers to an open-ended approach that aims to gain insight and understanding in a new or unexplored area (in contrast to \emph{confirmatory research}). It is often conducted when little is known about the phenomenon under study. It involves gathering information, identifying patterns, and formulating specific hypotheses for further investigation.



\emph{Hyperparameter tuning studies.} Aim to find the best-performing configuration for an ML model class, including baselines \citep{feurer2019hyperparameter,bischl2023hyperparameter}. 
Tuned models can then be compared more objectively and fairly. Hyperparameter tuning (or the lack of it) is an important source of variation in benchmark studies~\citep{bouthillier2021variance} and has been shown to have a strong effect on the outcome of results (see for example the references in \citealp{bouthillier2021variance} or our introduction). Treating the hyperparameter optimization as part of the problem of quantifying the performance of an algorithm has been suggested by \citet{bergstra2011tpe} and \citet{bergstra2013science}.

\emph{Insight-oriented exploratory research.} Refers to experimental research in ML that aims to gain insight, rather than inventing/developing a new method. It does not necessarily involve a very specific hypothesis to be pursued, but it is about improving the understanding and knowledge of a problem, a (class of) existing methods, or a phenomenon.  

\emph{Method-developing exploratory research.} Refers to experimental research in ML carried out in the process of developing a new ML method. This can include method comparison experiments, but in particular, it refers to exploration that takes place during the development process. This may include, for example, trying different method variants or specifying hyperparameter configurations and implementation details.  


\emph{Operationalization.} 
\citet[p.~256]{chang2004inventing} provides the following definition in his glossary: \lq\lq The process of giving operational meaning to a concept where there was none before. Operationalization may or may not involve the specification of explicit measurement methods." \emph{Operational meaning} refers to \lq\lq the meaning of a concept that is embodied in the physical operations whose description involves the concept." For a thorough discussion, see Chapter 4 (pp.~197--219).


\emph{Replicability (vs. reproducibility).} There is no consistent use of these terms in the broader literature \citep[for discussions, e.g., see][]{barbaTerminologiesReproducibleResearch2018, plesserReproducibilityVsReplicability2018, gundersen2021fundamental, pineau2021repro}. We use the term reproducibility in a narrow technical sense (see the glossary entry on \emph{computational reproducibility}). In contrast, replicability here means arriving at the same scientific conclusions in a broad sense. This terminology is in line with the \citet{national2019reproducibility}. In terms of the reliability of results, it means that replicability is more important than reproducibility. Note that \citet{drummondReplicabilityNotReproducibility2009}, for example, uses the terms the reverse way. 

\emph{Reproducibility (computational).} 
Means that the provided code technically achieves the same result on the provided data, and not that code, experimental design, or analysis are error-free and that we can qualitatively reach the same conclusions for the same general question under slightly different technical conditions. 
It is thus not a sufficient condition for replicability. Note that \citet{tatmanPracticalTaxonomyReproducibility2018} differentiate three levels of reproducibility.



\emph{Two historically distinct types of testing theory}. This refers to two approaches to statistical testing developed by Ronald Aylmer Fisher (1890--1962) on the one side and Jerzy Neyman (1894--1981) and Egon Sharpe Pearson (1895--1980) on the other. Only the former includes $p$-values and a single (null) hypothesis. The latter includes two hypotheses and hinges on statistical power and Type I and II errors \citep[p.~413]{schneider2015null}. More generally, Fisher's approach is \lq\lq [b]ased on the concept of a \lq hypothetical infinite population'," \lq\lq[r]oots in inductive philosophy" and \lq\lq [a]pplies to any single experiment (short run)," while Neyman-Pearson's approach is \lq\lq [b]ased on a clearly defined population," \lq\lq [r]oots in deductive philosophy," and \lq\lq [a]pplies only to ongoing, identical repetitions of an experiment, not to any single experiment (long run)" \citep[p.~415, Table~1]{schneider2015null}. 

\emph{Validity.} Note that there is no concise definition of the term. In psychology, internal and external validity are differentiated in particular. According to \citet[p.~297]{campellValidity1957} internal validity asks if \lq\lq in fact the experimental stimulus make some significant difference in this specific instance?"  External validity, on the other hand, asks \lq\lq to what populations, settings, and variables can this effect be generalized?" The former appears to be closely related to in-distribution generalization performance in ML, the latter to out-of-distribution generalization. In contrast, \emph{The Stanford Encyclopedia of Philosophy} states for experiments in physics \citep{franklinExp2023}: \lq\lq Physics, and natural science in general, is a reasonable enterprise based on \emph{valid} [emphasis added] experimental evidence, criticism, and rational discussion." Several strategies that may be used to validate observations are specified. These include the following: 1) \lq\lq Experimental checks and calibration, in which the experimental apparatus reproduces known phenomena"; 2) \lq\lq Reproducing artifacts that are known in advance to be present"; 3) \lq\lq Elimination of plausible sources of error and alternative explanations of the result"; 4) \lq\lq Using the results themselves to argue for their validity"; 5) \lq\lq Using an independently well-corroborated theory of the phenomena to explain the results"; 6) \lq\lq Using an apparatus based on a well-corroborated theory"; 7) \lq\lq Using statistical arguments." However, it is emphasized that \lq\lq [t]here are many experiments in which these strategies are applied, but whose results are later shown to be incorrect [...]. Experiment is fallible. Neither are these strategies exclusive or exhaustive. No single one of them, or fixed combination of them, guarantees the validity of an experimental result" \citep{franklinExp2023}.

\end{document}